\documentclass{article}
\usepackage{iclr2027_conference,times}

\usepackage{amsmath,amsfonts,bm}

\def\eqref#1{equation~\ref{#1}}
\def\1{\bm{1}}

\def\va{{\bm{a}}}

\def\vh{{\bm{h}}}

\def\vz{{\bm{z}}}

\DeclareMathAlphabet{\mathsfit}{\encodingdefault}{\sfdefault}{m}{sl}
\SetMathAlphabet{\mathsfit}{bold}{\encodingdefault}{\sfdefault}{bx}{n}

\def\gT{{\mathcal{T}}}

\newcommand{\E}{\mathbb{E}}

\usepackage{hyperref}
\usepackage{url}

\usepackage{amssymb}
\usepackage{multirow}
\usepackage{booktabs}
\usepackage{tabularx}
\usepackage{graphicx}
\usepackage{xspace}
\usepackage{enumitem}
\usepackage{amsmath}
\usepackage[table]{xcolor}
\usepackage{caption}
\renewcommand{\eqref}[1]{(\ref{#1})} % math_commands.tex defines \eqref as ``equation~\ref''
\newcommand{\method}[0]{\textsc{PICA-WAM}\xspace}
\newcommand{\picaicon}{\raisebox{-0.15em}{\includegraphics[height=1em]{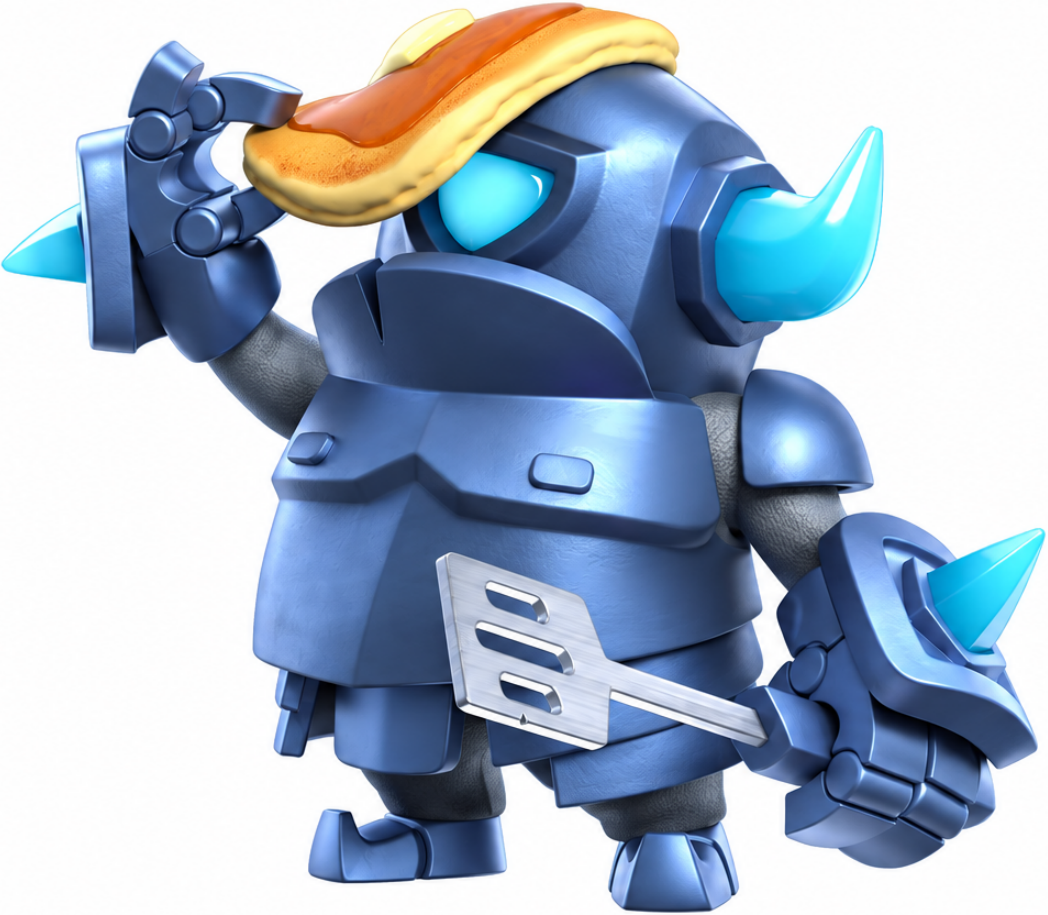}}}

\title{\picaicon \enspace PICA-WAM: Visual Perturbation Invariance via Cross-Appearance Self-Alignment in World Action Models}

\author{
Yude Zou$^{1,2}$\thanks{Equal contribution. \quad \textsuperscript{$\dagger$}Corresponding author.} \quad Yifei Yao$^{2}$\footnotemark[1] \quad Guanbo Jiao$^{2,3}$\footnotemark[1] \quad Jianming Ma$^{1,2}$ \quad Guanjie Zheng$^{1}$ \quad Xing Gao$^{2}$\textsuperscript{$\dagger$} \\
{\normalfont $^1$Shanghai Jiao Tong University \quad $^2$Shanghai Artificial Intelligence Laboratory \quad $^3$Fudan University}
}

\iclrfinalcopy
\begin{document}

\maketitle
\lhead{Preprint}
\begin{figure}[h!]
\centering
\includegraphics[width=1\textwidth]{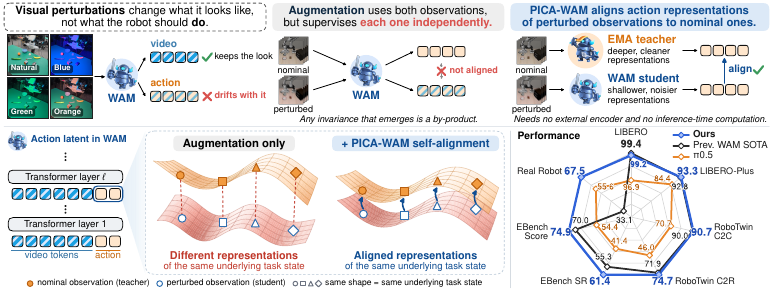}
\caption{\textbf{Overview of \method.} Top: augmentation supervises each observation independently, whereas \method aligns shallower-layer action representations of the student on the perturbed observation to deeper-layer ones of its EMA teacher on the nominal observation. Bottom left: schematic of action representations. Matching shapes denote the same underlying task state across nominal and perturbed observations. Bottom right: performance. Ours is the best \method result, and each axis is scaled linearly to the range of results per benchmark.}
\label{fig:teaser}
\end{figure}
\begin{abstract}
% \vspace{-5pt}
World Action Models (WAMs) jointly predict future video and generate actions, yet remain sensitive to task-irrelevant visual perturbations, \emph{i.e.}, changes in visual appearance that should not affect robot behavior. We identify a gap in conventional WAM training: video prediction encourages representations to preserve appearance information, while action generation does not explicitly enforce invariance to task-irrelevant appearance. 
% World Action Models (WAMs) jointly learn to predict future video and to generate actions, yet they remain brittle under visual perturbations: task-irrelevant changes in how a scene is observed, which alter what it \emph{looks like} but not what the robot \emph{should do}.
%We study a gap in conventional WAM training: the video prediction objective requires faithful reconstruction of appearance details, while consistent action generation under task-irrelevant appearance changes is not explicitly enforced.
Data augmentation exposes WAMs to such perturbations but does not require perturbed and nominal observations to produce consistent action representations, and alignment with external pretrained features does not guarantee the desired invariance. 
%Training on augmented observations exposes a WAM to such changes but never ties the representation of a perturbed observation to that of its nominal counterpart, and aligning to external pretrained encoders imports features that are not themselves invariant to task-irrelevant appearance.
To address this gap, we propose PICA-WAM, a self-supervised training framework that casts perturbation invariance as an auxiliary cross-appearance self-alignment task for WAMs. 
An exponential-moving-average (EMA) teacher processes the nominal observation to provide a deeper-layer target, while the online WAM processes a perturbed observation and aligns its shallower-layer action representation to this target through a lightweight projection head.
%An exponential-moving-average copy of the WAM reads the nominal observation at a slightly lower noise level and provides a later-layer target. The online WAM reads a perturbed view of the same state and aligns its earlier-layer action representation to this target through a lightweight projection head.
Across four strong WAMs evaluated on RoboTwin 2.0, LIBERO-Plus, EBench, and a real robot, PICA-WAM consistently improves robustness under visual perturbations, preserves nominal performance in simulation, and raises the real robot success rate. PICA-WAM further reduces the sensitivity of video features to task-irrelevant appearance and its propagation into actions, as shown by paired layer-wise diagnostics. Code: \url{https://github.com/YudeZou/PICA-WAM-release}.

\end{abstract}
\vspace{-10pt}
\section{Introduction}
\label{sec:intro}

World Action Models (WAMs) build robot policies on top of pretrained video generators, coupling predictions of how a scene will evolve with the actions that drive it~\citep{yuan2026fastwam,li2026lingbotva,ye2026dreamzero,wang2026openwam,cai2026internw0}. Video--action co-training transfers the dynamics prior of large-scale video pretraining into control, and WAMs now approach saturation on standard manipulation benchmarks such as LIBERO~\citep{liu2023libero} and RoboTwin~\citep{mu2024robotwin,mu2025robotwin,chen2025robotwin}. Yet these models remain brittle when observations depart from the training distribution~\citep{hong2026steering}. On LIBERO-Plus~\citep{fei2025liberoplus} and RoboTwin 2.0 Clean-to-Random~\citep{yuan2026qwen}, shifts in appearance, viewpoint, sensing, and scene configuration can substantially reduce success. The task goal remains the same, but the policy must act under unfamiliar observations and starting conditions.

We view this brittleness as a gap in the training objectives of WAMs. The video prediction objective requires faithful reconstruction of appearance details such as lighting and texture, while reliable control calls for consistent actions when the same underlying task state is observed under different task-irrelevant appearance variations, and joint video--action training does not explicitly enforce this consistency. As the action expert reads video-stream representations, task-irrelevant appearance variation leaks into the action pathway.

Existing approaches address this gap only partially. \emph{(i) Data augmentation}~\citep{tobin2017domain,kostrikov2021drq} exposes the model to perturbed observations but never requires their representations to agree with those of their nominal counterparts. The resulting invariance is incidental rather than enforced: in our layer-wise analysis, augmentation lowers appearance sensitivity in intermediate video layers, but sensitivity rises above the unaugmented baseline in layers 21--25 (Figure~\ref{fig:video_stream_and_bands}, left).
\emph{(ii) External representation alignment} regularizes WAM features toward priors from outside the model, such as semantic vision encoders~\citep{qiu2026agra,yan2026robustwam,wang2026stwam}, 3D trajectory fields~\citep{yang2026fourdwam}, 4D tracking features~\citep{miao2026internw0}, or rendered Gaussian fields~\citep{zhang2026gaussianwam}. These priors provide semantic and geometric guidance, but alignment to an external teacher does not keep the representations of the same underlying task state consistent under visual perturbations. This leaves a need for explicit cross-appearance supervision.
\emph{(iii) Self-alignment} constructs representation targets from the generative model itself, without an external encoder. In image and video generation, \citet{jiang2025sra} and \citet{chefer2026selfflow} align shallower, noisier features to deeper, cleaner features of an exponential-moving-average (EMA) teacher of the model~\citep{tarvainen2017mean,grill2020byol}, matching the gains of alignment to external encoders~\citep{yu2025repa}.
These methods align differently noised versions of the same sample, without matching nominal and perturbed observations of the same underlying task state. Self-alignment also remains underexplored in WAMs, where joint video--action prediction raises the question of whether to align the video stream, the action stream, or both.

To this end, we propose \textbf{\method}, a self-supervised training framework that promotes visual \textbf{P}erturbation \textbf{I}nvariance via an auxiliary \textbf{C}ross-\textbf{A}ppearance self-alignment task in \textbf{WAM}s. An EMA teacher of the WAM reads the nominal observation at a lower noise level and provides a deeper-layer target. The online WAM reads a perturbed observation of the same underlying task state and aligns its shallower-layer action representation to this target through a lightweight projection head (Figure~\ref{fig:method}). Each pair thus spans three gaps. The \emph{appearance gap} supplies the invariance signal. The \emph{noise-level} and \emph{layer} gaps, which \citet{jiang2025sra} identified as key ingredients of self-alignment, make the target stronger than the student it supervises. We apply the objective to the action stream, which directly encourages consistency in the representation used for control while leaving the video prediction objective unchanged. In our ablations, this choice outperforms aligning the video stream or both streams. The teacher and the head are discarded after training, so \method adds no parameters and no computation at inference and applies to any WAM with a layered backbone.

We evaluate \method on four strong WAM backbones across simulation and real-world settings: Fast-WAM~\citep{yuan2026fastwam} on RoboTwin 2.0 Clean-to-Random and LIBERO-Plus, LingBot-VA~\citep{li2026lingbotva} on RoboTwin 2.0 Clean-to-Random, InternW0-$\Delta$~\citep{miao2026internw0} on RoboTwin 2.0 Clean-to-Random and LIBERO-Plus, and OpenWAM-$\alpha$~\citep{wang2026openwam} on EBench~\citep{gao2026ebench} and a real robot. \method consistently improves robustness over the corresponding baselines and achieves the best results among the compared methods on all three simulation benchmarks: built on InternW0-$\Delta$, it reaches 74.7\% on RoboTwin 2.0 Clean-to-Random and 93.3\% on LIBERO-Plus, and built on OpenWAM-$\alpha$, it reaches 61.4\% on EBench. Even without embodied pretraining, it brings Fast-WAM to success rates comparable to those of models with large-scale embodied pretraining. Paired analyses further show that it reduces both the appearance sensitivity of video features and its propagation into actions, including under perturbation families unseen during training.

In summary, our main contributions are as follows:
\begin{itemize}[leftmargin=*,itemsep=2pt,topsep=2pt]
    \item We propose \method, a self-supervised training framework for WAMs in which the action representation from a perturbed observation is aligned to a cleaner, deeper target computed from the nominal observation by an EMA teacher of the model, and that needs no external encoder, no additional data, and no inference-time computation.
    \item We design a paired diagnostic protocol that traces a visual perturbation from layer-wise features to executed actions, and show that augmentation alone leaves late video layers appearance-sensitive and lets this sensitivity reach the action, whereas \method reduces both.
    \item We demonstrate consistent robustness gains on RoboTwin 2.0 Clean-to-Random, LIBERO-Plus, EBench, and real-world manipulation under visual perturbation across four strong WAMs with and without embodied pretraining, while preserving in-distribution performance.
\end{itemize}

\section{Related Work}
\label{sec:related}

\paragraph{World action models.}
Early work synthesizes future frames and infers actions from them~\citep{du2023unipi}, or conditions a policy on video pretraining and predictive representations~\citep{wu2024gr1,hu2025vpp}. Later models jointly denoise video and actions~\citep{zhu2025uwm,liao2025genie,kim2026cosmospolicy}, and current WAMs scale this recipe with stronger video backbones, action experts, and embodied pretraining~\citep{yuan2026fastwam,li2026lingbotva,wang2026openwam,ye2026dreamzero,cai2026internw0,miao2026internw0,chen2026abot,chen2026lawam,guo2026unified}.
Yet WAMs remain brittle under visual perturbations~\citep{fei2025liberoplus,yuan2026qwen}, and their predicted futures can hallucinate training-domain content~\citep{wang2026stwam}.
Remedies include inference-time activation steering~\citep{hong2026steering} and training-time priors from outside the model: semantic features from pretrained encoders such as DINOv3 and V-JEPA or from text~\citep{qiu2026agra,yan2026robustwam,wang2026stwam,liu2026cswam,feng2026copper,he2026sgwam}, geometric supervision such as 3D trajectory fields, 4D tracking features, and Gaussian fields~\citep{ma2026geosemwam,yang2026fourdwam,miao2026internw0,zhang2026gaussianwam}, and distractor-invariant object slots~\citep{liu2026oawam}. Data augmentation and domain randomization~\citep{tobin2017domain,kostrikov2021drq} expose a policy to appearance variation but supervise each observation independently. Self-Guided WAM~\citep{zhao2026sgwamself} uses an EMA teacher, but to supervise future dynamics rather than cross-appearance agreement.
None of these methods constrains WAM representations across task-irrelevant appearance variations of the same underlying task state. \method instead casts perturbation invariance as a cross-appearance self-alignment task, without any external encoder.

\paragraph{Representation alignment in generative models.}
REPA~\citep{yu2025repa} aligns intermediate diffusion-transformer features with a pretrained visual encoder to accelerate training and improve generation, and follow-up work extends it to end-to-end VAE tuning~\citep{leng2025repae} and video generation~\citep{zhang2025videorepa}. Several of the semantic WAM methods above follow the same recipe. Such targets inherit the properties of the external encoder: frozen pretrained encoders, including DINO- and SigLIP-style ones, are strongly appearance-coupled, and the action information decodable from their features collapses under noise and blur~\citep{yeom2026actionrelevant}.
Self-alignment instead constructs targets within the model. SRA~\citep{jiang2025sra} aligns shallower-layer features of a student at a higher noise level with deeper-layer features of an EMA teacher at a lower noise level. Self-Flow~\citep{chefer2026selfflow} noises the student's tokens at heterogeneous levels while the teacher observes the cleaner input, scaling self-alignment to image, video, and audio generation. These approaches train a student to agree with a slowly updated teacher on a different view of the same input, but their views differ in noise level or network depth, not in task-irrelevant appearance, so the agreement does not encode appearance invariance.
\method makes task-irrelevant appearance the difference between the two views: the teacher reads the nominal observation and the student a perturbed observation of the same underlying task state.

\section{Method}
\label{sec:method}

\begin{figure}[t]
\centering
\includegraphics[width=1\textwidth]{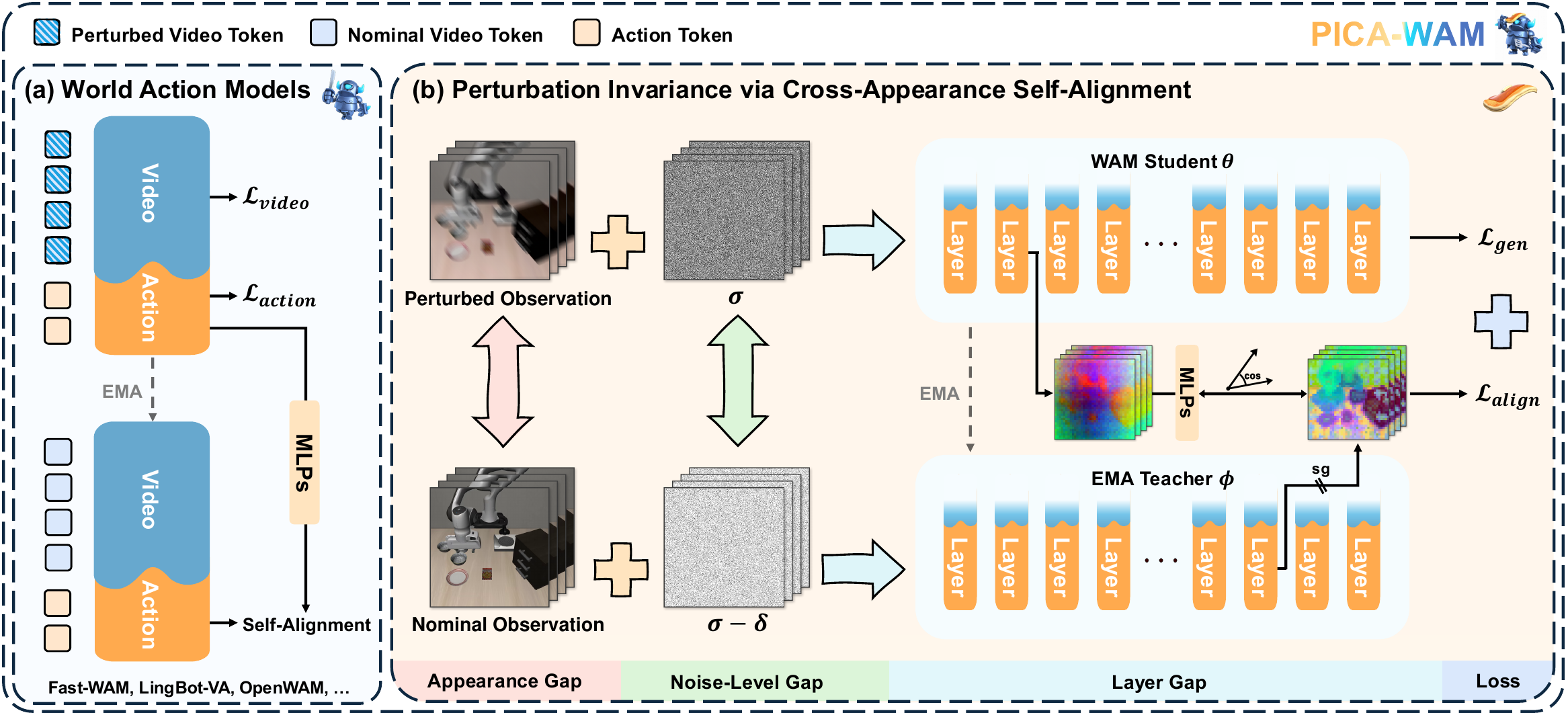}
\caption{\textbf{Framework of \method.}
\textbf{(a)} A two-stream WAM is trained with $\mathcal{L}_{\text{gen}}=\mathcal{L}_{\text{video}}+\mathcal{L}_{\text{action}}$. \method adds an auxiliary self-alignment task between the action streams of the model and its EMA teacher.
\textbf{(b)} The student $\theta$ reads a perturbed observation at video noise level $\sigma$ and is trained with $\mathcal{L}_{\text{gen}}$. The EMA teacher $\phi$ reads the nominal observation of the same underlying task state at $\sigma-\delta$. Projected shallower-layer action representations of the student are aligned to deeper-layer ones of the teacher under stop-gradient (sg), giving $\mathcal{L}_{\text{align}}$.}
\label{fig:method}
\vspace{-10pt}
\end{figure}

\subsection{Preliminaries: World Action Models}
\label{sec:prelim}

A WAM takes the current observation and a language instruction $c$ and jointly predicts future video and a chunk of actions. We consider the common two-stream design~\citep{yuan2026fastwam,li2026lingbotva,wang2026openwam} (Figure~\ref{fig:method}a): a video stream initialized from a pretrained video diffusion transformer and an action stream are coupled layer by layer through joint attention, in the style of a mixture of transformers. Let $\vz$ be the latent video clip, whose first frame encodes the current observation, and let $\va$ be the action chunk. The two streams are trained jointly under a flow-matching objective~\citep{lipman2023flow,liu2023rectified}. The future frames are noised as $\vz_\sigma = (1-\sigma)\vz + \sigma\bm{\epsilon}$ while the first frame is kept clean, and the actions as $\va_\tau = (1-\tau)\va + \tau\bm{\eta}$, with $\bm{\epsilon},\bm{\eta}\sim\mathcal{N}(\bm{0},\bm{I})$. The model $\theta$ predicts the velocities $v$ of both streams, and the generative loss is the sum of a video term and an action term:
\begin{equation}
\begin{aligned}
\mathcal{L}_{\text{gen}} &= \mathcal{L}_{\text{video}} + \mathcal{L}_{\text{action}} \\
&= \E\Big[\big\|v^{\text{vid}}_\theta(\vz_\sigma,\va_\tau,c) - (\bm{\epsilon}-\vz)\big\|^2 + \big\|v^{\text{act}}_\theta(\vz_\sigma,\va_\tau,c) - (\bm{\eta}-\va)\big\|^2\Big],
\end{aligned}
\label{eq:gen}
\end{equation}
where the video term is taken over the future frames. We write $\vh^{(l)}(\vz_\sigma,\va_\tau;\theta)\in\mathbb{R}^{K\times d}$ for the hidden states of the $K$ action tokens at layer $l$ of the action stream. Because action tokens attend to video tokens at every layer, $\vh^{(l)}$ depends on the observation and predicted future through the video stream, and so does any task-irrelevant appearance variation in it. Unless stated otherwise, ``appearance'' below refers to this task-irrelevant visual variation.

\subsection{Action Invariance to Visual Perturbations}
\label{sec:formulation}

Let $\vz$ be a nominal clip as recorded and $T\in\gT$ a visual perturbation, such as a photometric, sensor, or mild geometric change, that alters how the scene is observed but not the underlying task state. Applying $T$ to every frame yields a perturbed clip $\vz'=T(\vz)$ for which the demonstrated actions $\va$ remain correct. We seek action representations that are invariant to such perturbations:
\begin{equation}
\vh^{(l)}(\vz'_\sigma,\va_\tau;\theta) \approx \vh^{(l)}(\vz_\sigma,\va_\tau;\theta), \qquad \forall\, T\in\gT .
\label{eq:invariance}
\end{equation}
We impose this property on the action stream only. The video stream must retain appearance to predict the future frames of $\vz'$, so requiring invariance of the whole representation would conflict with the video objective. The action stream is the representation from which actions are decoded, and constraining it still shapes the video representations it reads through joint attention (Section~\ref{sec:exp_diag}).

Training on perturbed clips, i.e., minimizing $\E_{T}\big[\mathcal{L}_{\text{gen}}(T(\vz),\va)\big]$, is equivalent to augmentation-only training, and it does not impose Eq.~\eqref{eq:invariance}. Specifically, in this objective, each clip is a separate training example, and the loss scores only the predictions for its future frames and its actions. It never scores the intermediate representations, and it never considers a perturbed clip together with the nominal clip of the same underlying task state. Whether the two clips are represented alike is therefore not a training target, and any invariance that emerges is a by-product. Two properties of WAMs work against this by-product. First, the video objective requires faithful reconstruction of the perturbed future frames, so the video stream must encode the appearance introduced by $T$. Second, the action tokens read these video representations through joint attention at every layer, so this appearance variation enters the action stream. The action objective penalizes this variation only when it changes the predicted actions on the training clips and never removes it from the intermediate representations. Section~\ref{sec:exp_diag} examines this layer by layer. Imposing Eq.~\eqref{eq:invariance} thus requires supervision on the representation itself that relates each perturbed clip to its nominal counterpart. We provide it by computing, from the nominal clip, a target for $\vh^{(l)}(\vz'_\sigma,\va_\tau;\theta)$, as described next.

\subsection{Self-Alignment across Appearances}
\label{sec:pica}

We obtain the target through \emph{cross-appearance self-alignment} (Figure~\ref{fig:method}b): the WAM computes a target from the nominal clip, and its representation of the perturbed clip is aligned to that target.

\textbf{Student and teacher.} The student is the online model $\theta$. It reads the perturbed clip at video noise level $\sigma$, and its layer-$m$ action representation is mapped by a lightweight MLP projection head $g_\psi$. The teacher is an EMA of the model, $\phi\leftarrow\mu\phi+(1-\mu)\theta$. It reads the nominal clip at a lower video noise level $\sigma-\delta$ with $\delta>0$, and provides the target action representation from a deeper layer $n>m$. Both branches differ only in the observed clip and its future video noise level. The alignment loss is the negative cosine similarity per action token, where $\mathrm{sg}[\cdot]$ denotes stop-gradient:
\vspace{-5pt}
\begin{equation}
\mathcal{L}_{\text{align}} = -\frac{1}{K}\sum_{k=1}^{K}\cos\!\Big(g_\psi\big(\vh^{(m)}_k(\vz'_\sigma,\va_\tau;\theta)\big),\ \mathrm{sg}\big[\vh^{(n)}_k(\vz_{\sigma-\delta},\va_\tau;\phi)\big]\Big).
\label{eq:align}
\vspace{-5pt}
\end{equation}
\textbf{Three gaps.} Each pair spans three gaps between student and teacher (bottom of Figure~\ref{fig:method}b). The \emph{appearance gap} ($\vz'$ versus $\vz$) supplies the invariance signal. Student and teacher share the underlying task state, the noisy actions, and the instruction, and their inputs differ in appearance only through $T$. Matching the teacher target therefore requires the action representation of the student to discard what $T$ changes, which is the invariance of Eq.~\eqref{eq:invariance}. The alignment thus provides the supervision relating a perturbed clip to its nominal counterpart that augmentation-only training lacks (Section~\ref{sec:formulation}). Since only the student receives gradients, the representation of the perturbed clip is pulled toward that of the nominal clip and not the reverse, which matches deployment, where the model receives perturbed observations. The \emph{noise-level} and \emph{layer gaps} make the target stronger than the student's own representation: the teacher reads a less noisy input and provides its representation from a deeper layer, both of which \citet{jiang2025sra} found to yield better targets for alignment.

\textbf{Training and inference.} The overall objective is:
\begin{equation}
\mathcal{L} = \mathcal{L}_{\text{gen}}(\vz',\va) + \lambda\,\mathcal{L}_{\text{align}},
\label{eq:total}
\end{equation}
optimized over $\theta$ and $\psi$, with the teacher updated by EMA after each step. The alignment term adds one gradient-free teacher forward pass per step, which can stop at layer $n$. At inference, the teacher and the projection head are discarded, so \method leaves the architecture, the parameter count, and the sampling procedure of the WAM unchanged. The perturbation families, layer choices, and other hyperparameters are listed in Appendix~\ref{app:impl}.

\section{Experiments}
\label{sec:exp}

\subsection{Experimental Setup}
\label{sec:exp_setup}

\textbf{Simulation Benchmarks.} RoboTwin 2.0 Clean-to-Random~\citep{chen2025robotwin,yuan2026qwen} trains on 50 clean demonstrations for each of 50 bimanual tasks, recorded with a fixed background, default lighting, no distractors, and a fixed table height. Each task is evaluated for 100 episodes in this clean setting and in a domain-randomized setting that randomizes background textures, lighting, distractor clutter, and table height and uses unseen language instructions. LIBERO-Plus~\citep{fei2025liberoplus} evaluates policies trained on the four LIBERO suites~\citep{liu2023libero} on 10,030 perturbed tasks along seven dimensions: robot initial state, language instruction, lighting, background texture, sensor noise, object layout, and camera viewpoint. EBench~\citep{gao2026ebench} evaluates policies on 26 tasks: 19 mobile manipulation tasks and 7 table-top dexterous-and-precise tasks in household, retail, and industrial scenes. Policies are evaluated through the official online platform, in which the episodes of each task vary the objects, the background, the instruction, or a mix of them.

\textbf{Backbones and variants.} We apply \method to four WAMs. Fast-WAM~\citep{yuan2026fastwam}, which has no embodied pretraining, is evaluated on RoboTwin 2.0 and LIBERO. The other three are embodied-pretrained: LingBot-VA~\citep{li2026lingbotva} on RoboTwin 2.0, InternW0-$\Delta$~\citep{miao2026internw0} on RoboTwin 2.0 and LIBERO, and OpenWAM-$\alpha$~\citep{wang2026openwam} on EBench and a real robot (Section~\ref{sec:exp_real}). We use Fast-WAM in its joint mode, in which the video stream takes a single denoising step before the action is generated from the resulting video features. On LIBERO, we further train three controlled variants of Fast-WAM with the same data, schedule, and inference procedure: Baseline, trained on nominal clips, +Aug, trained on perturbed clips, and \method, which adds the alignment term to +Aug. More details are provided in Appendix~\ref{app:impl}.

\subsection{Does Augmentation Enforce Invariance?}
\label{sec:exp_diag}

To test the argument of Section~\ref{sec:formulation} that augmentation-only training leaves intermediate representations unconstrained, we design a paired diagnostic protocol. It compares the three controlled Fast-WAM variants (Section~\ref{sec:exp_setup}), which differ only in whether they are trained on perturbed clips and whether they are aligned. Each perturbed LIBERO-Plus observation is paired with the nominal observation of the same simulator state under the same noise seed, so that the two inputs differ only in how the scene is observed. The protocol traces the perturbation from features to executed actions at three levels. (i) \emph{Layer-wise sensitivity} is the feature distance between the two observations divided by the distance between the nominal observations of two different states in the same episode. Raw feature distances differ in scale across layers and models, and normalizing by a change in task state removes this difference, so that a value of 1 means that the perturbation changes the features as much as a change in task state does. (ii) A \emph{layer-band intervention} replaces the video features of one layer band in a nominal prediction with those from the perturbed observation, which tests whether the sensitivity remaining in that band shifts the action. (iii) \emph{Trajectory drift} is the distance between the end-effector positions reached by executing the action chunks predicted from the two observations. We report separately the \emph{seen} perturbation families, whose type is contained in $T$ (Gaussian, motion, and zoom blur, and fog), and the \emph{unseen} ones (glass blur, lighting, background, and camera viewpoint on LIBERO-Plus). All diagnostics use the single-step video denoising of deployment. Appendix~\ref{app:diag} details how the pairs are constructed and how each metric is computed.

\begin{figure}[t]
\centering
\includegraphics[width=1\textwidth]{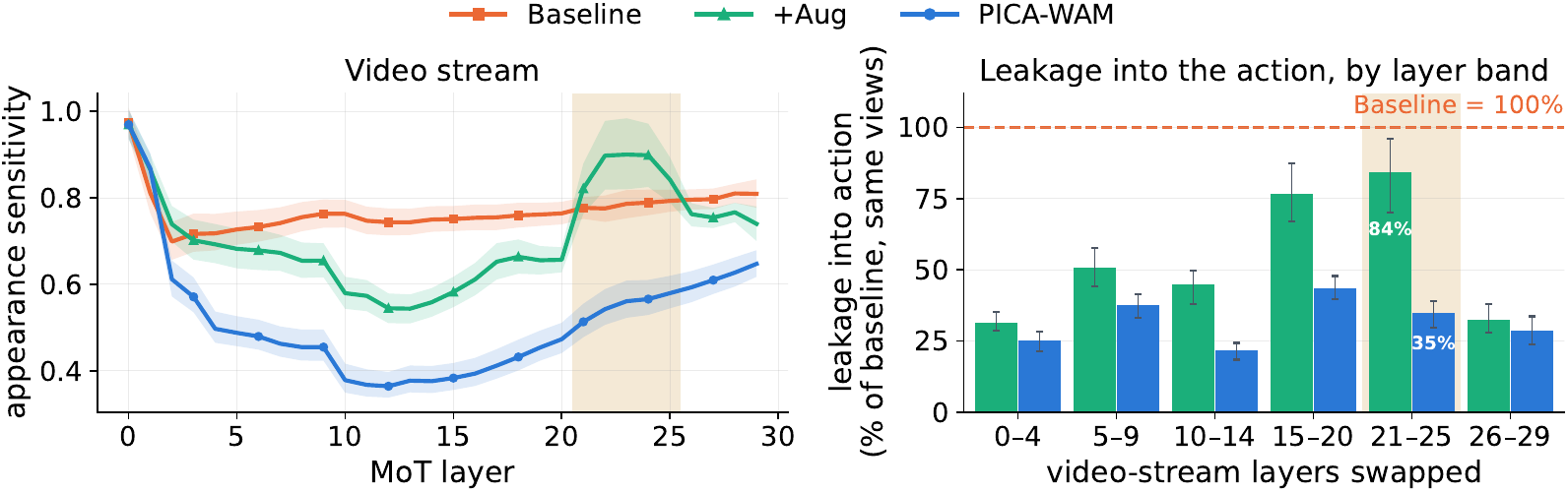}
\vspace{-15pt}
\caption{\textbf{Video-stream appearance sensitivity and its transfer into actions.}
Left: layer-wise sensitivity of video features on unseen perturbation families (Section~\ref{sec:exp_diag}). Right: action deviation caused by swapping the perturbed video features of one layer band into a nominal prediction, as a median ratio to that of the Baseline on the same observation pairs. In the highlighted layers 21--25, the sensitivity of +Aug rebounds, and these layers retain 84\% of the action leakage of the Baseline under +Aug, compared with 35\% under \method.}
\label{fig:video_stream_and_bands}
\vspace{-5pt}
\end{figure}

\begin{table}[t]
\centering
\scriptsize
\setlength{\tabcolsep}{4pt}
\renewcommand{\arraystretch}{1.15}
\caption{\textbf{Paired Consistency under LIBERO-Plus Visual Perturbations.}
Sensitivity is measured at layer 8 of each stream, and drift is the end-effector distance after executing the 32-step action chunks predicted from the paired observations. Lower is better. Metrics and seen/unseen families are defined in Section~\ref{sec:exp_diag}. Bold and underline denote the best and second-best values.}
\vspace{-5pt}
\label{tab:appearance_consistency}
\begin{tabularx}{\linewidth}{l*{6}{>{\centering\arraybackslash}X}}
\toprule
 & \multicolumn{2}{c}{\textbf{Video Sensitivity} $\downarrow$}
 & \multicolumn{2}{c}{\textbf{Action Sensitivity} $\downarrow$}
 & \multicolumn{2}{c}{\textbf{Drift (mm)} $\downarrow$} \\
\cmidrule(lr){2-3}\cmidrule(lr){4-5}\cmidrule(lr){6-7}
\textbf{Method} & \textbf{Seen} & \textbf{Unseen}
 & \textbf{Seen} & \textbf{Unseen}
 & \textbf{Seen} & \textbf{Unseen} \\
\midrule
\textbf{Baseline} & 0.610 & 0.755 & 0.407 & 0.486 & 16.14 & 16.42 \\
\rowcolor{black!4}
\textbf{+ Aug} & \underline{0.530} & \underline{0.655} & \underline{0.146} & \underline{0.189} & \underline{9.41} & \underline{11.24} \\
\midrule
\textbf{PICA-WAM (Ours)} & \textbf{0.369} & \textbf{0.455} & \textbf{0.057} & \textbf{0.075} & \textbf{5.49} & \textbf{6.21} \\
\bottomrule
\end{tabularx}
\vspace{-5pt}
\end{table}

\begin{table}[t]
\centering
\scriptsize
\setlength{\tabcolsep}{7pt}
\renewcommand{\arraystretch}{1.15}
\caption{\textbf{Evaluation Results on RoboTwin 2.0 Clean-to-Random.}
Method groups distinguish models with and without large-scale embodied pretraining.
$^{\dagger}$, $^{\ddagger}$, and $^{\S}$ denote methods built on the Fast-WAM, LingBot-VA, and InternW0-$\Delta$ backbones, respectively.
Bold and underline mark the best and second-best values within each group.}
\vspace{-5pt}
\label{tab:robotwin}
\begin{tabularx}{\linewidth}{l*{3}{>{\centering\arraybackslash}X}}
\toprule
\textbf{Method} & \textbf{Clean} & \textbf{Random} & \textbf{Avg.} \\
\midrule
\rowcolor{black!8}
\multicolumn{4}{c}{\textit{\textbf{Method without Emb. PT.}}} \\
\midrule
\textbf{StarVLA}~\citep{ye2026starvla}
 & 46.5 & 3.2 & 24.8 \\
\rowcolor{black!4}
\textbf{AHA-WAM}~\citep{cai2026ahawam}
 & 64.3 & 3.2 & 33.8 \\
\textbf{Fast-WAM}~\citep{yuan2026fastwam}
 & 77.8 & 1.9 & 39.9 \\
\rowcolor{black!4}
\textbf{Copper-Policy}~\citep{feng2026copper}
 & 70.8 & \underline{13.0} & 41.9 \\
\textbf{CSWAM}$^{\dagger}$~\citep{liu2026cswam}
 & \underline{78.5} & 9.4 & \underline{43.9} \\
\midrule
\textbf{PICA-WAM (Ours)}$^{\dagger}$
 & \textbf{88.0} & \textbf{39.6} & \textbf{63.8} \\
\midrule
\rowcolor{black!8}
\multicolumn{4}{c}{\textit{\textbf{Method with Emb. PT.}}} \\
\midrule
\textbf{GR00T-N1.7}~\citep{nvidia2025gr00tn1}
 & 43.6 & 20.7 & 32.2 \\
\rowcolor{black!4}
\textbf{GalaxeaVLA}~\citep{jiang2025galaxea}
 & 62.7 & 12.7 & 37.7 \\
\textbf{Xiaomi-Robotics-0}~\citep{cai2026xiaomi}
 & 62.9 & 18.2 & 40.6 \\
\rowcolor{black!4}
\textbf{EventVLA}~\citep{yang2026eventvla}
 & 65.6 & 15.7 & 40.7 \\
\textbf{ABot-M0}~\citep{yang2026abot}
 & 57.4 & 30.4 & 43.9 \\
\rowcolor{black!4}
\textbf{X-VLA}~\citep{zheng2026x}
 & 68.0 & 20.9 & 44.5 \\
\textbf{Spatial Forcing}~\citep{li2025spatialforcing}
 & 77.2 & 26.7 & 52.0 \\
\rowcolor{black!4}
\textbf{\boldmath $\pi_{0.5}$}~\citep{intelligence2025pi_}
 & 70.7 & 46.0 & 58.4 \\
\textbf{GigaBrain-0.7}~\citep{team2026gigabrain}
 & 66.8 & 67.9 & 67.4 \\
\rowcolor{black!4}
\textbf{Qwen-RobotManip}~\citep{yuan2026qwen}
 & 84.7 & 69.4 & 77.1 \\
\textbf{X-WAM}~\citep{guo2026unified}
 & 70.0 & 25.8 & 47.9 \\
\rowcolor{black!4}
\textbf{LingBot-VA}~\citep{li2026lingbotva}
 & 81.2 & 29.8 & 55.5 \\
\textbf{Robust-WAM}$^{\ddagger}$~\citep{yan2026robustwam}
 & 81.2 & 34.4 & 57.8 \\
\rowcolor{black!4}
\textbf{OLA-Geo}~\citep{robotwin2026leaderboard}
 & 83.0 & 33.6 & 58.3 \\
\textbf{4D-WAM}$^{\ddagger}$~\citep{yang2026fourdwam}
 & 81.5 & 41.8 & 61.7 \\
\rowcolor{black!4}
\textbf{WorldScape Policy 2.0}~\citep{su2026worldscape}
 & 84.9 & 38.5 & 61.7 \\
\textbf{CSWAM}~\citep{liu2026cswam}
 & 84.5 & 45.2 & 64.8 \\
\rowcolor{black!4}
\textbf{OpenWAM-$\alpha$}~\citep{wang2026openwam}
 & 89.4 & 48.7 & 69.0 \\
\textbf{InternW0-$\Delta$}~\citep{miao2026internw0}
 & \underline{90.0} & \underline{71.9} & \underline{81.0} \\
\midrule
\textbf{PICA-WAM (Ours)}$^{\ddagger}$
 & 83.1 & 57.0 & 70.1 \\
\textbf{PICA-WAM (Ours)}$^{\S}$
 & \textbf{90.7} & \textbf{74.7} & \textbf{82.7} \\
\bottomrule
\end{tabularx}
\vspace{-5pt}
\end{table}

\textbf{Augmentation leaves late video layers appearance-sensitive.}
Figure~\ref{fig:video_stream_and_bands} (left) shows the video-stream sensitivity on unseen families. Relative to the Baseline, +Aug lowers the sensitivity in the intermediate layers, but the sensitivity rebounds in layers 21--25 and exceeds that of the Baseline, so augmentation does not remove appearance variation from the video features that the action stream reads in these layers. \method is the least sensitive from layer 2, although it aligns only the action stream: constraining the action stream also shapes the video features through joint attention.

\textbf{Residual sensitivity reaches the action.}
Figure~\ref{fig:video_stream_and_bands} (right) reports the layer-band intervention, normalized by the action deviation of the Baseline on the same observation pairs. Under +Aug, bands 15--20 and 21--25 carry the largest leakage, and layers 21--25, where the sensitivity rebounds, retain 84\% of the leakage of the Baseline. The appearance variation remaining in these features therefore still shifts the action. \method lowers the leakage in every band, to 35\% in layers 21--25.

\textbf{Alignment reduces sensitivity on seen and unseen families.}
Table~\ref{tab:appearance_consistency} reports the sensitivity of both streams at a shallower layer (layer 8). On both groups, +Aug lowers the video sensitivity of the Baseline only moderately, from 0.610 to 0.530 on seen and from 0.755 to 0.655 on unseen families, whereas \method lowers it to 0.369 and 0.455. In the action stream, where the action objective penalizes appearance variation that changes the predicted actions, +Aug lowers the sensitivity of the Baseline from 0.407 to 0.146 and from 0.486 to 0.189, and \method further lowers it to 0.057 and 0.075. These reductions over +Aug are of similar size on the two groups, so the effect of the alignment is not confined to the perturbation types contained in $T$.

\textbf{Lower sensitivity carries over to executed actions.}
The drift in Table~\ref{tab:appearance_consistency} follows the same pattern. +Aug reduces the drift of the Baseline from 16.14 to 9.41\,mm on seen families but only from 16.42 to 11.24\,mm on unseen families, i.e., by 42\% versus 32\%. \method reduces it to 5.49 and 6.21\,mm, by 66\% and 62\%, respectively. Appendix~\ref{app:diag} provides further analyses and visualizations, including the executed trajectories and video prediction fidelity.

\subsection{Simulation Benchmarks}
\label{sec:exp_sim}

\begin{table}[t]
\centering
\scriptsize
\setlength{\tabcolsep}{3pt}
\renewcommand{\arraystretch}{1.15}
\caption{\textbf{Evaluation Results on LIBERO and LIBERO-Plus.}
Method groups distinguish models with and without large-scale embodied pretraining before LIBERO adaptation.
$^{\dagger}$ and $^{\S}$ denote methods built on the Fast-WAM and InternW0-$\Delta$ backbones, respectively.
Bold denotes the best value and underline the second best within each group.}
\label{tab:main_libero_plus}
\stepcounter{footnote}% marks the results we evaluated ourselves; the text is \footnotetext in the LIBERO paragraph of Section 4.3
\vspace{-5pt}
\begin{tabularx}{\linewidth}{l*{9}{>{\centering\arraybackslash}X}}
\toprule
 & & \multicolumn{8}{c}{\textbf{LIBERO-Plus}} \\
\cmidrule(lr){3-10}
\textbf{Method} & \textbf{LIBERO}
 & \textbf{Robot} & \textbf{Language} & \textbf{Light} & \textbf{BG}
 & \textbf{Noise} & \textbf{Layout} & \textbf{Camera} & \textbf{Avg.} \\
\midrule
\rowcolor{black!8}
\multicolumn{10}{c}{\textit{\textbf{Method without Emb. PT.}}} \\
\midrule
\textbf{Fast-WAM}~\citep{yuan2026fastwam}
 & 97.6 & 44.5 & 68.9 & 78.2 & 53.7 & 37.7 & 60.7 & 16.4 & 51.5 \\
\rowcolor{black!4}
\textbf{Robust-WAM}$^{\dagger}$~\citep{yan2026robustwam}
 & 97.9 & 43.4 & 73.6 & 90.2 & 62.6 & 56.8 & 67.8 & 28.7 & 58.9 \\
\textbf{4D-WAM}$^{\dagger}$~\citep{yang2026fourdwam}
 & \underline{98.6} & 64.3 & \textbf{90.6} & \underline{94.3} & 57.7 & 69.1 & 79.2 & 45.2 & 71.0 \\
\rowcolor{black!4}
\textbf{GaussianWAM}$^{\dagger}$~\citep{zhang2026gaussianwam}
 & 97.6 & 61.6 & 75.4 & 90.8 & 58.2 & 83.0 & 76.7 & 54.7 & 71.3 \\
\textbf{ST-WAM}~\citep{wang2026stwam}
 & \textbf{98.7} & 60.1 & 79.3 & 93.0 & 74.2 & 79.5 & 74.3 & 55.4 & 72.8 \\
\rowcolor{black!4}
\textbf{Self-Guided WAM}~\citep{zhao2026sgwamself}
 & 98.5 & 48.9 & 81.4 & 89.8 & \underline{86.1} & 80.7 & 74.2 & \underline{58.6} & 73.0 \\
\textbf{SG-WAM}~\citep{he2026sgwam}
 & \textbf{98.7} & \underline{78.3} & 81.7 & \textbf{96.5} & \textbf{88.9} & \underline{90.7} & \textbf{85.3} & 54.8 & \underline{81.3} \\
\midrule
\textbf{PICA-WAM (Ours)}$^{\dagger}$
 & 98.2 & \textbf{95.4} & \underline{86.5} & 94.1 & 79.3 & \textbf{94.6} & \underline{82.6} & \textbf{64.7} & \textbf{85.2} \\
\midrule
\rowcolor{black!8}
\multicolumn{10}{c}{\textit{\textbf{Method with Emb. PT.}}} \\
\midrule
\textbf{\boldmath $\pi_0$}~\citep{black2024pi_0}
 & 94.4 & 6.0 & 58.8 & 85.0 & 81.4 & 79.0 & 68.9 & 13.8 & 53.6 \\
\rowcolor{black!4}
\textbf{OpenVLA-OFT}~\citep{kim2025fine}
 & 97.1 & 31.9 & 79.5 & 88.7 & 93.3 & 75.8 & 74.2 & 56.4 & 69.6 \\
\textbf{X-VLA}~\citep{zheng2026x}
 & 98.1 & 89.7 & 75.7 & 88.2 & 96.0 & 62.7 & 71.8 & 23.4 & 71.4 \\
\rowcolor{black!4}
\textbf{\boldmath $\pi_{0.5}$}~\citep{intelligence2025pi_}
 & 96.9 & 73.6 & 80.8 & 96.2 & 94.1 & 89.0 & 84.5 & 78.4 & 84.4 \\
\textbf{InternVLA-A1.5}~\citep{ma2026internvlaa15}
 & 98.9 & 55.1 & 86.9 & 96.4 & 98.2 & 95.6 & 85.2 & 83.1 & 84.8 \\
\rowcolor{black!4}
\textbf{Qwen-RobotManip}~\citep{yuan2026qwen}
 & 99.2 & 83.9 & 86.5 & \textbf{98.6} & \textbf{99.9} & \textbf{97.9} & \underline{87.5} & \underline{89.9} & 91.4 \\
\textbf{OpenWAM-$\alpha$}~\citep{wang2026openwam}
 & \underline{99.3} & 76.1 & 88.0 & 97.0 & 87.1 & 39.8 & 77.5 & 33.8 & 69.2 \\
\rowcolor{black!4}
\textbf{LaWAM}\footnotemark[\value{footnote}]~\citep{chen2026lawam}
 & 98.6 & \textbf{96.5} & 85.2 & 93.6 & 94.0 & 68.3 & 83.5 & 33.8 & 77.7 \\
\textbf{LingBot-VA}\footnotemark[\value{footnote}]~\citep{li2026lingbotva}
 & 98.5 & \underline{93.9} & 84.1 & 92.3 & 80.1 & 75.6 & 79.3 & 62.9 & 80.7 \\
\rowcolor{black!4}
\textbf{ABot-M0.5}~\citep{chen2026abot}
 & \textbf{99.4} & 87.4 & 88.6 & 94.0 & 89.7 & 75.5 & 85.2 & 70.5 & 83.4 \\
\textbf{InternW0-$\Delta$}\footnotemark[\value{footnote}]~\citep{miao2026internw0}
 & 99.0 & 91.1 & \underline{92.9} & 95.8 & \underline{99.4} & 94.0 & \textbf{88.4} & \textbf{90.6} & \underline{92.8} \\
\midrule
\textbf{PICA-WAM (Ours)}$^{\S}$
 & 99.2 & 91.7 & \textbf{95.3} & \underline{98.0} & 99.3 & \underline{96.6} & \textbf{88.4} & 86.7 & \textbf{93.3} \\
\bottomrule
\end{tabularx}
\vspace{-5pt}
\end{table}

\begin{table}[t]
\centering
\scriptsize
\setlength{\tabcolsep}{3pt}
\renewcommand{\arraystretch}{1.15}
\caption{\textbf{Evaluation Results on EBench}.
Mobile Manipulation contains 19 tasks and Table-Top Dexterous-and-Precise contains 7 tasks, and all averages weight tasks equally.
$^{\dagger}$ denotes a method built on the OpenWAM-$\alpha$ backbone.
Results of other methods are taken from the EBench leaderboard.
Bold and underline mark the best and second-best values.}
\vspace{-5pt}
\label{tab:ebench}
\begin{tabularx}{\linewidth}{l*{6}{>{\centering\arraybackslash}X}}
\toprule
 & \multicolumn{2}{c}{\textbf{Mobile Manipulation}} & \multicolumn{2}{c}{\textbf{Table-Top Dex. \& Prec.}} & \multicolumn{2}{c}{\textbf{Overall}} \\
\cmidrule(lr){2-3}\cmidrule(lr){4-5}\cmidrule(lr){6-7}
\textbf{Method} & \textbf{SR} & \textbf{Score} & \textbf{SR} & \textbf{Score} & \textbf{SR} & \textbf{Score} \\
\midrule
\textbf{X-VLA}~\citep{zheng2026x}
 & 29.3 & 40.2 & 8.6 & 24.1 & 23.7 & 35.8 \\
\rowcolor{black!4}
\textbf{Fast-WAM}~\citep{yuan2026fastwam}
 & 34.0 & 46.1 & 2.9 & 13.0 & 25.6 & 37.1 \\
\textbf{Cosmos3-Edge}~\citep{agarwal2026cosmos}
 & 33.2 & 46.6 & 18.6 & 31.6 & 29.3 & 42.5 \\
\rowcolor{black!4}
\textbf{GigaBrain-0.7}~\citep{team2026gigabrain}
 & 31.6 & 41.4 & 37.9 & 59.0 & 33.3 & 46.0 \\
\textbf{\boldmath $\pi_0$}~\citep{black2024pi_0}
 & 34.0 & 46.8 & 32.9 & 49.4 & 33.7 & 47.5 \\
\rowcolor{black!4}
\textbf{InternVLA-A1.5}~\citep{ma2026internvlaa15}
 & 42.3 & 56.2 & 12.1 & 21.6 & 34.2 & 46.8 \\
\textbf{\boldmath $\pi_{0.5}$}~\citep{intelligence2025pi_}
 & 45.9 & 57.8 & 29.3 & 45.3 & 41.4 & 54.4 \\
\rowcolor{black!4}
\textbf{Qwen-RobotManip}~\citep{yuan2026qwen}
 & 43.9 & 57.6 & \textbf{50.0} & \textbf{69.7} & 45.6 & 60.8 \\
\textbf{GPT-6-Astra-ICL}~\citep{yao2026frontier}
 & 56.6 & 73.1 & 20.0 & 44.7 & 46.7 & 65.4 \\
\rowcolor{black!4}
\textbf{Amapbot}
 & 49.3 & 64.2 & \underline{47.9} & 63.0 & 48.9 & 63.9 \\
\textbf{InternW0-$\Delta$}~\citep{miao2026internw0}
 & 55.0 & 70.0 & 33.6 & 55.2 & 49.2 & 66.0 \\
\rowcolor{black!4}
\textbf{Vela}~\citep{li2026vela}
 & 51.7 & 66.1 & 44.3 & 64.1 & 49.7 & 65.5 \\
\textbf{OpenWAM-$\alpha$}~\citep{wang2026openwam}
 & \underline{60.2} & \underline{74.1} & 42.1 & 59.6 & \underline{55.3} & \underline{70.0} \\
\midrule
\textbf{PICA-WAM (Ours)}$^{\dagger}$
 & \textbf{67.5} & \textbf{78.3} & 45.0 & \underline{66.3} & \textbf{61.4} & \textbf{74.9} \\
\bottomrule
\end{tabularx}
\vspace{-5pt}
\end{table}

\textbf{RoboTwin 2.0 Clean-to-Random.}
Table~\ref{tab:robotwin} reports the results. Built on LingBot-VA, \method raises the success rate from 29.8\% to 57.0\% in the randomized setting and outperforms Robust-WAM and 4D-WAM, which use the same backbone but align to external encoders or 3D representations. Based on InternW0-$\Delta$, \method raises the success rate from 71.9\% to 74.7\% in the randomized setting and from 90.0\% to 90.7\% in the clean setting, achieving the highest average success rate (82.7\%) among all compared methods. Without embodied pretraining, \method achieves 88.0\% in the clean setting and 39.6\% in the randomized setting, the highest in this group.

\textbf{LIBERO and LIBERO-Plus.}
Table~\ref{tab:main_libero_plus} reports the results.\footnotetext[\value{footnote}]{In Table~\ref{tab:main_libero_plus}, the LIBERO result of InternW0-$\Delta$ is evaluated by us with its official checkpoint, and the LIBERO-Plus results of LaWAM and LingBot-VA are reproduced by us.} Based on InternW0-$\Delta$, \method raises the average success rate on LIBERO-Plus from 92.8\% to 93.3\%, the highest among all compared methods. The gains mainly come from language (+2.4), sensor noise (+2.6), and lighting (+2.2). Without embodied pretraining, \method reaches an average success rate of 85.2\% on LIBERO-Plus, exceeding all compared VLAs and WAMs with large-scale embodied pretraining except InternW0-$\Delta$ and Qwen-RobotManip, while retaining 98.2\% on LIBERO. 

\textbf{EBench.}
Table~\ref{tab:ebench} reports the results. Based on OpenWAM-$\alpha$, \method raises the overall success rate from 55.3\% to 61.4\% and the score from 70.0 to 74.9, achieving the highest overall success rate and score among the compared methods. The gains hold in both task categories: +7.3 SR and +4.2 score on Mobile Manipulation, where it achieves the highest SR and score among all compared methods, and +2.9 SR and +6.7 score on Table-Top Dexterous-and-Precise.

% \textcolor{blue}{\textbf{Complementarity with external representation alignment.}}
% \textcolor{blue}{InternW0-$\Delta$ already aligns its video expert to an external teacher: during pretraining, it distills 4D geometry and motion descriptors from a frozen Track4World tracker into the video expert~\citep{miao2026internw0}. \method still improves its average success rate on LIBERO-Plus (+0.5) and RoboTwin 2.0 Clean-to-Random (+1.7), with the largest gains in the randomized setting of RoboTwin 2.0 (+2.8) and under sensor noise (+2.6) and lighting (+2.2) on LIBERO-Plus. The 4D teacher transfers geometric and motion priors, but it does not require the nominal and perturbed views of the same state to be represented alike, which is the supervision that \method adds. The same holds on LingBot-VA, where \method outperforms Robust-WAM and 4D-WAM, which align to external encoders or 3D representations.}

\subsection{Ablation Studies}
\label{sec:ablation}

\begin{table}[t]
\centering
\scriptsize
\setlength{\tabcolsep}{5pt}
\renewcommand{\arraystretch}{1.15}
\caption{\textbf{Ablation Results on LIBERO-Plus.}
We report success rates (\%) and their unweighted mean on the five dimensions that perturb the visual scene. Robot initial state and language are excluded because they perturb the initial pose and the instruction. Baseline and +Aug are the controlled variants of Section~\ref{sec:exp_setup}. + SRA adds self-alignment with the noise-level and layer gaps but without the appearance gap. The \method rows add the alignment of Eq.~\eqref{eq:align} to +Aug and differ in the aligned stream. Bold and underline denote the best and second-best values.}
\label{tab:libero_plus_ablation}
\vspace{-5pt}
\begin{tabularx}{\linewidth}{l*{6}{>{\centering\arraybackslash}X}}
\toprule
\textbf{Method} & \textbf{Light} & \textbf{BG} & \textbf{Noise} & \textbf{Layout} & \textbf{Camera} & \textbf{Avg.} \\
\midrule
\textbf{Baseline}
 & 93.2 & 60.6 & 67.2 & 81.4 & 45.7 & 69.6 \\
\rowcolor{black!4}
\textbf{\quad+ SRA}~\citep{jiang2025sra}
 & \underline{94.1} & 60.6 & 70.1 & \underline{82.5} & 50.8 & 71.6 \\
\midrule
\textbf{+ Aug}
 & 93.3 & 74.2 & 87.1 & 78.3 & 62.1 & 79.0 \\
\rowcolor{black!4}
\textbf{\quad+ SRA}
 & 92.7 & \textbf{80.0} & 92.1 & 79.0 & \underline{62.7} & 81.3 \\
\textbf{\quad+ PICA-WAM\textsubscript{align video}}
 & 94.0 & 79.0 & 92.8 & 82.2 & 60.0 & \underline{81.6} \\
\rowcolor{black!4}
\textbf{\quad+ PICA-WAM\textsubscript{align both}}
 & \textbf{95.4} & 76.3 & \underline{93.6} & 79.8 & 58.7 & 80.7 \\
\midrule
\textbf{\quad+ PICA-WAM\textsubscript{align action} (Ours)}
 & \underline{94.1} & \underline{79.3} & \textbf{94.6} & \textbf{82.6} & \textbf{64.7} & \textbf{83.1} \\
\bottomrule
\end{tabularx}
\vspace{-5pt}
\end{table}

Table~\ref{tab:libero_plus_ablation} ablates the alignment on LIBERO-Plus. Since LIBERO-Plus applies each perturbation dimension separately, it shows how each design choice affects robustness to each type of perturbation. SRA-style self-alignment, which keeps the noise-level and layer gaps but removes the appearance gap, raises the average success rate of the Baseline from 69.6\% to 71.6\% and that of +Aug from 79.0\% to 81.3\%. \method raises +Aug to 83.1\%, with the largest gains over +Aug on sensor noise (+7.5), background (+5.1), and object layout (+4.3). The three \method variants differ only in the aligned stream. Aligning the action stream performs best (83.1\%), ahead of aligning the video stream (81.6\%) or both streams (80.7\%). This is consistent with Section~\ref{sec:formulation}: aligning the video stream also constrains the representations that predict future frames, whereas aligning only the action stream leaves the video stream free to encode the appearance that video prediction requires.

\subsection{Real-World Experiments}
\label{sec:exp_real}

\begin{figure}[!t]
\centering
\includegraphics[width=1\textwidth]{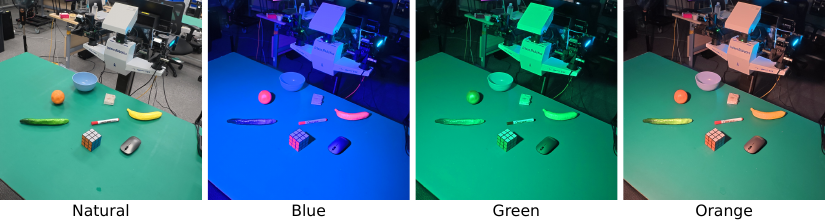}
\vspace{-15pt}
\caption{\textbf{Real-world evaluation scene under the four lighting conditions.}
The photos show the eight task objects on the evaluation tabletop, which differs from the one used for the demonstrations. The bowl, the mouse, and the pen are new instances. The demonstrations are recorded under natural lighting, whereas the blue, green, and orange lights shift the apparent colors of the tabletop and the objects. For example, the banana appears pink under blue light.}
\label{fig:real_world}
\vspace{-5pt}
\end{figure}

We evaluate on an ARX-Lift2 dual-arm robot with one head camera and two wrist cameras. The eight tasks require the left arm to pick up a banana, a bowl, a cucumber, a computer mouse, an orange, a pen, a Rubik's cube, or a tissue, with 25 demonstrations per task. Every trial uses visual perturbations absent from the demonstrations: a new tabletop texture and new instances of the bowl, the mouse, and the pen with different appearances. We evaluate under four lighting conditions: the natural lighting of the demonstrations and three colored lights (blue, green, and orange), as shown in Figure~\ref{fig:real_world}. For each lighting condition, we run five trials per task with a different object layout in each trial, and a trial succeeds if the robot picks up the object. We compare \method based on OpenWAM-$\alpha$ with OpenWAM-$\alpha$ and $\pi_{0.5}$, all fine-tuned on the same demonstrations.

\begin{table}[t]
\centering
\scriptsize
\setlength{\tabcolsep}{2pt}
\renewcommand{\arraystretch}{1.15}
\caption{\textbf{Real-World Success Rates per Task} (\%).
Each task is evaluated in 20 trials, five under each lighting condition. Avg.\ is over all 160 trials.
$^{*}$ marks objects replaced with new instances at evaluation.
$^{\dagger}$ denotes a method built on the OpenWAM-$\alpha$ backbone. Bold and underline mark the best and second-best values.}
\vspace{-5pt}
\label{tab:real_world}
\begin{tabularx}{\linewidth}{l*{2}{>{\hsize=0.9625\hsize\centering\arraybackslash}X}>{\hsize=1.3\hsize\centering\arraybackslash}X*{6}{>{\hsize=0.9625\hsize\centering\arraybackslash}X}}
\toprule
\textbf{Method} & \textbf{Banana} & \textbf{Bowl}$^{*}$ & \textbf{Cucumber} & \textbf{Mouse}$^{*}$ & \textbf{Orange} & \textbf{Pen}$^{*}$ & \textbf{Cube} & \textbf{Tissue} & \textbf{Avg.} \\
\midrule
\textbf{\boldmath $\pi_{0.5}$}
 & \underline{85.0} & \textbf{100.0} & \underline{85.0} & 0.0 & 0.0 & \textbf{100.0} & 0.0 & \textbf{75.0} & \underline{55.6} \\
\rowcolor{black!4}
\textbf{OpenWAM-$\alpha$}
 & 5.0 & \underline{95.0} & 5.0 & \underline{25.0} & 0.0 & \underline{70.0} & 0.0 & \underline{65.0} & 33.1 \\
\midrule
\textbf{PICA-WAM (Ours)}$^{\dagger}$
 & \textbf{95.0} & \textbf{100.0} & \textbf{90.0} & \textbf{30.0} & \textbf{75.0} & \textbf{100.0} & \textbf{10.0} & 40.0 & \textbf{67.5} \\
\bottomrule
\end{tabularx}
\vspace{-5pt}
\end{table}

As shown in Table~\ref{tab:real_world}, \method raises the success rate of OpenWAM-$\alpha$ from 33.1\% to 67.5\% and exceeds $\pi_{0.5}$. It improves over OpenWAM-$\alpha$ on seven of the eight tasks, most on the banana, cucumber, and orange tasks, where OpenWAM-$\alpha$ succeeds in at most 5\% of the trials and \mbox{\method} in at least 75\%. $\pi_{0.5}$ either succeeds in most trials of a task or fails in all of them, and it fails in every trial of the mouse, orange, and Rubik's cube tasks, whereas \method succeeds on every task. It is also the only method that succeeds on the orange and Rubik's cube tasks. 

\section{Conclusion}
\label{sec:conclusion}
We introduced \method, a self-supervised training framework that casts perturbation invariance as an auxiliary cross-appearance self-alignment task for WAMs. An EMA teacher of the model computes a target from the nominal observation at a lower noise level and a deeper layer, and the action representation of a perturbed observation is aligned to this target. The teacher and the head are discarded after training. Across Fast-WAM, LingBot-VA, InternW0-$\Delta$, and OpenWAM-$\alpha$, \method improves success rates under visual perturbations on RoboTwin 2.0 Clean-to-Random, LIBERO-Plus, EBench, and a real robot. Paired diagnostics show that it reduces the appearance sensitivity of both streams and the resulting trajectory drift, including on unseen perturbation families.

% \subsection*{AI use statement}
%
% In this work, we used generative AI tools to aid and polish the writing and for retrieval and discovery, such as identifying related work. Specifically, large language models were used to edit the text for readability, clarity, and consistency, and to search for related work. We have not used generative AI tools for the other tasks with required disclosure, such as proposing hypotheses, designing experiments, implementing methods, or interpreting results. We have reviewed all AI-assisted work and take responsibility for the final content of this work.

\bibliography{iclr2027_conference}
\bibliographystyle{iclr2027_conference}

\appendix
\section{Implementation Details}
\label{app:impl}

\paragraph{Perturbation distribution.}
$T$ is applied to each camera independently, and its parameters are sampled once per clip and shared by all frames of that camera. For each camera, $T$ applies photometric transforms only, additive noise only, or both, with probabilities $1/4$, $1/4$, and $1/2$. The photometric transforms scale brightness and contrast by factors in $[0.7, 1.3]$ and saturation by a factor in $[0.75, 1.25]$, shift hue by up to $0.04$, apply a gamma in $[0.8, 1.25]$, and change exposure by up to $\pm 0.25$ stops. The additive noise is Gaussian with standard deviation $0.015$ for pixel values in $[0,1]$, and the same noise field is applied to all frames. With probability $0.25$, $T$ also crops the clip to a side-length fraction in $[0.92, 1.0]$ and rotates it by up to $\pm 8^\circ$. On LIBERO, and for InternW0-$\Delta$ also on RoboTwin 2.0, with probability $0.5$, $T$ further applies one optical corruption drawn from isotropic blur ($\sigma \in [0.4, 5]$ pixels), directional blur (length in $[3, 21]$ pixels), zoom blur (maximum zoom in $[1.02, 1.2]$), and fog (strength in $[0.3, 2.5]$). The teacher always reads the unperturbed clip.

\paragraph{Alignment.}
Table~\ref{tab:align_hparams} lists the alignment hyperparameters, which are shared by the four backbones. The projection head $g_\psi$ is a two-layer MLP with a GELU activation that maps the action-stream width to a hidden width of 512 and back.

\begin{table}[h]
\centering
\small
\renewcommand{\arraystretch}{1.15}
\caption{\textbf{Alignment hyperparameters} shared by Fast-WAM, LingBot-VA, InternW0-$\Delta$, and OpenWAM-$\alpha$. Layers are indexed from 0.}
\label{tab:align_hparams}
\begin{tabular}{ll}
\toprule
\textbf{Hyperparameter} & \textbf{Value} \\
\midrule
Aligned stream & action \\
Student layer $m$ / teacher layer $n$ & 8 / 20 \\
Loss weight $\lambda$ & 0.05 \\
Noise-level gap $\delta$ & $\mathcal{U}[0, 0.2)$ \\
Projection head & Linear--GELU--Linear, hidden width 512 \\
EMA decay & 0.9999 (fp32) \\
\bottomrule
\end{tabular}
\end{table}

\paragraph{Training.}
Table~\ref{tab:train_hparams} lists the training configurations. InternW0-$\Delta$ and OpenWAM-$\alpha$ are fine-tuned from their pretrained checkpoints, following the post-training configuration of InternW0-$\Delta$ and the EBench training configuration of OpenWAM-$\alpha$, and $\pi_{0.5}$ is fine-tuned from its base checkpoint with an action horizon of 32. On RoboTwin 2.0, all backbones are trained on the 50 clean demonstrations of each task, and LingBot-VA takes full episodes as training samples. On LIBERO, the three Fast-WAM variants are evaluated at the same checkpoint and differ only in $T$ and $\lambda$.

\begin{table}[h]
\centering
\small
\renewcommand{\arraystretch}{1.15}
\caption{\textbf{Training configurations.} InternW0-$\Delta$ and OpenWAM-$\alpha$ are trained on 33-frame clips, at a stride of 4 from three cameras for OpenWAM-$\alpha$. LingBot-VA's batch size counts episodes.}
\label{tab:train_hparams}
\begin{tabular}{llrr}
\toprule
\textbf{Benchmark} & \textbf{Model} & \textbf{Steps} & \textbf{Batch} \\
\midrule
\multirow{3}{*}{RoboTwin 2.0} & LingBot-VA & 30{,}000 & 16 \\
 & Fast-WAM & 21{,}145 & 128 \\
 & InternW0-$\Delta$ & 5{,}370 & 1{,}024 \\
\midrule
\multirow{2}{*}{LIBERO} & Fast-WAM & 21{,}700 & 128 \\
 & InternW0-$\Delta$ & 16{,}275 & 256 \\
\midrule
EBench & OpenWAM-$\alpha$ & 100{,}000 & 256 \\
\midrule
\multirow{2}{*}{Real robot} & OpenWAM-$\alpha$ & 2{,}000 & 128 \\
 & $\pi_{0.5}$ & 2{,}000 & 128 \\
\bottomrule
\end{tabular}
\end{table}

\paragraph{Inference.}
On RoboTwin 2.0, Fast-WAM runs in its joint mode, whereas LingBot-VA uses its default inference settings. On LIBERO and LIBERO-Plus, Fast-WAM also runs in its joint mode: the video stream takes a single Euler step from $\sigma=0.9998$ to $\sigma=0.953$, and the action stream then runs 10 denoising steps while attending to the video features at that noise level. The diagnostics in Section~\ref{sec:exp_diag} use the same procedure. InternW0-$\Delta$ on RoboTwin 2.0, LIBERO, and LIBERO-Plus and OpenWAM-$\alpha$ on EBench and the real robot use their default inference settings.

\section{Diagnostic Protocol}
\label{app:diag}

\paragraph{Paired observations.}
We replay LIBERO demonstrations in the LIBERO-Plus base environment and record the full simulator state at five control steps per episode. For the lighting, background, and camera families, we load the corresponding LIBERO-Plus scene, which differs from the base scene only in its scene description or camera pose, and set it to the recorded state. We verify that the resulting simulator state and end-effector pose match the nominal ones. We use one variant per LIBERO-Plus difficulty level for each of these families. For the sensor-noise families, we apply the LIBERO-Plus corruption functions to the nominal render at five severities. All observations of a state share the same noise seed.

\paragraph{Feature sensitivity.}
For each layer, we compute the relative $\ell_2$ distance between the perturbed and the nominal features, and divide it by the same distance between the nominal features of the state and those of a different state in the same episode, at least eight control steps apart. The video features are the hidden states from which each layer computes its video keys and values, restricted to the observed-frame tokens. The action features are the action-token inputs of each layer at the first denoising step, where the action input is the same noise for both observations. We pool distances over perturbed observations and report 95\% bootstrap confidence intervals over states.

\paragraph{Layer-band intervention.}
We partition the 30 video layers into the bands 0--4, 5--9, 10--14, 15--20, 21--25, and 26--29. For each band, we replace the video hidden states of the nominal observation with those of the perturbed observation, recompute the keys and values of these layers, and denoise the action again with the same noise. We report the resulting action deviation from the nominal action, normalized by that of the Baseline on the same observation pair, and compare \method with +Aug by a paired Wilcoxon test.

\begin{figure}[t]
\centering
\includegraphics[width=1\textwidth]{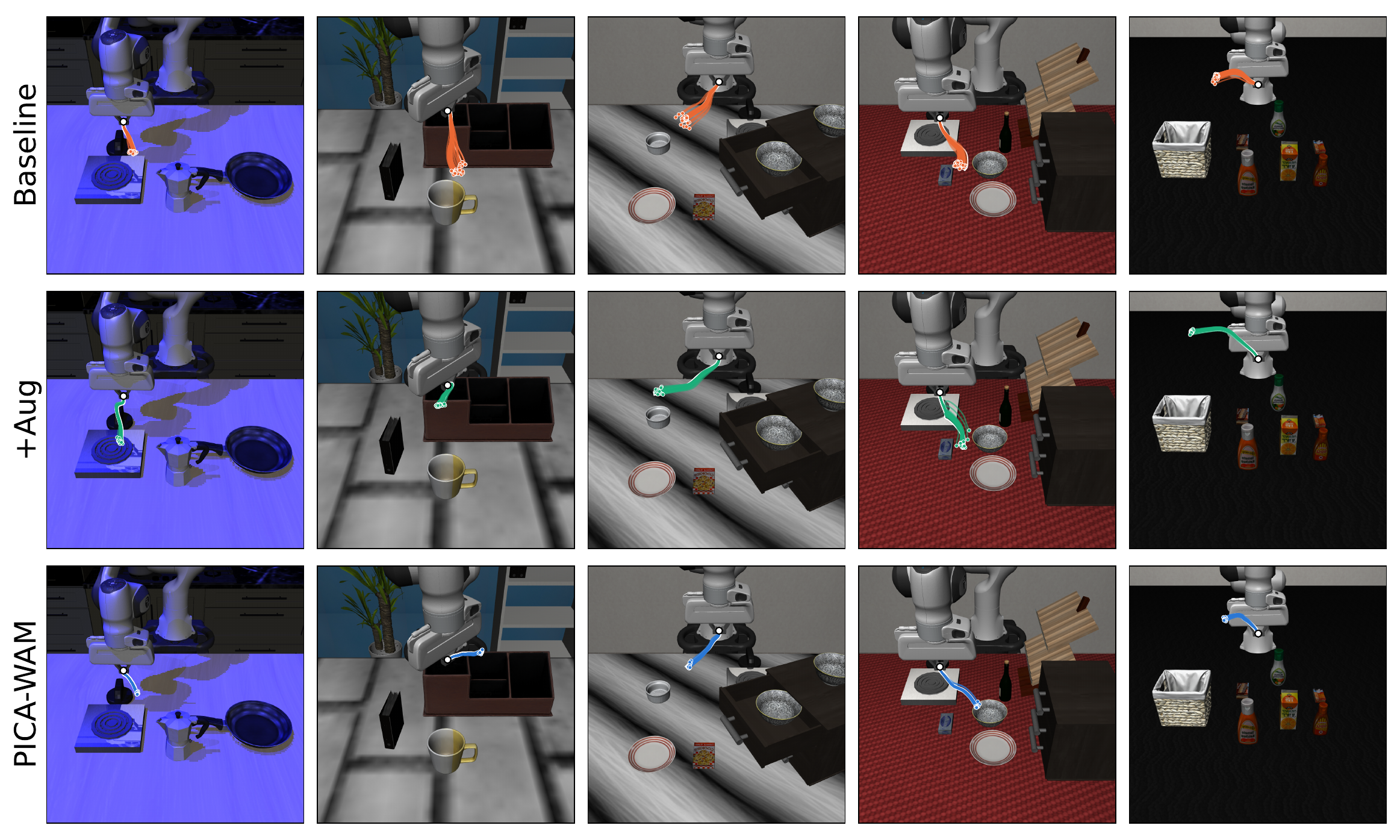}
\caption{\textbf{Executed trajectories under visual perturbations.}
Each column shows one LIBERO-Plus state under a perturbed observation. Each curve is the end-effector path obtained by executing the 32-step action chunk predicted from a different perturbed observation of the same state with the same noise seed. The white dot marks the start position. A tighter fan indicates more consistent actions across appearances.
The first three columns are the states with the largest baseline drift. The last two are drawn at random.}
\label{fig:fanout}
\end{figure}

\paragraph{Trajectory drift.}
Each predicted 32-step action chunk is executed open-loop from the recorded state in the base scene, which the perturbations do not change physically. The drift is the distance between end-effector positions reached from the nominal and the perturbed observation. On every family, \method reduces the drift of +Aug. Figure~\ref{fig:fanout} shows the trajectories for five states.

\paragraph{Video prediction fidelity.}
We draw 96 training clips, construct their nominal and perturbed observations with $T$, run the full video denoising schedule, and compare the predicted future frames with the ground truth of the same observation. Table~\ref{tab:video_fidelity} shows that \method and +Aug have the same fidelity on both observations. Both are below the Baseline on nominal observations and above it on perturbed observations.

\begin{table}[t]
\centering
\small
\setlength{\tabcolsep}{6pt}
\renewcommand{\arraystretch}{1.15}
\caption{\textbf{Video prediction fidelity} on 96 LIBERO training clips, averaged over the third-person and wrist cameras, with 95\% bootstrap confidence intervals. The perturbed observations are constructed with the training distribution $T$.}
\label{tab:video_fidelity}
\begin{tabular}{lcccc}
\toprule
 & \multicolumn{2}{c}{\textbf{Nominal observation}} & \multicolumn{2}{c}{\textbf{Perturbed observation}} \\
\cmidrule(lr){2-3}\cmidrule(lr){4-5}
\textbf{Method} & \textbf{PSNR} $\uparrow$ & \textbf{SSIM} $\uparrow$ & \textbf{PSNR} $\uparrow$ & \textbf{SSIM} $\uparrow$ \\
\midrule
Baseline & 24.62 & 0.864 & 19.82 & 0.734 \\
+ Aug & 22.38 & 0.829 & 24.14 & 0.813 \\
\method & 22.44 & 0.828 & 24.12 & 0.813 \\
\bottomrule
\end{tabular}
\end{table}

\end{document}